\documentclass[conference]{IEEEtran}

\usepackage{amsmath,amssymb}
\usepackage{booktabs}
\usepackage{multirow}
\usepackage{array}
\usepackage{graphicx}
\usepackage{url}
\usepackage{cite}
\usepackage{stfloats}
\usepackage{placeins}
\usepackage{microtype}

\newcommand{\ours}{IRFE-ECG}

\begin{document}

\title{Separating Expert Retention from Autonomous Source Inference in Raw-ECG-Replay-Free Continual ECG Deployment}

\author{
\makebox[\textwidth][c]{%
\begin{minipage}[t]{0.30\textwidth}
\centering
{\normalfont\bfseries 1\textsuperscript{st} Yufan Lu}\\
{\itshape Xidian University}\\
24019100468@stu.xidian.edu.cn
\end{minipage}
\hfill
\begin{minipage}[t]{0.30\textwidth}
\centering
{\normalfont\bfseries 2\textsuperscript{nd} Xinhui Liu}\\
{\itshape Xidian University}\\
24049200220@stu.xidian.edu.cn
\end{minipage}
\hfill
\begin{minipage}[t]{0.30\textwidth}
\centering
{\normalfont\bfseries 3\textsuperscript{rd} Chenyang Xu}\\
{\itshape Xidian University}\\
xcy@ieee.org
\end{minipage}
}

\\[1.2em]

\makebox[\textwidth][c]{%
\begin{minipage}[t]{0.46\textwidth}
\centering
{\normalfont\bfseries 4\textsuperscript{th} Yuxi Zhou\textsuperscript{*}}\\
{\itshape Department of Computer Science, Tianjin University of Technology}\\
{\itshape DCST, BNRist, RIIT, Institute of Internet Industry, Tsinghua University}\\
joy\_yuxi@pku.edu.cn
\end{minipage}
\hfill
\begin{minipage}[t]{0.40\textwidth}
\centering
{\normalfont\bfseries 5\textsuperscript{th} Hao Wang\textsuperscript{*}}\\
{\itshape Xidian University}\\
haow@ieee.org\\
\textsuperscript{*}Corresponding authors: Yuxi Zhou and Hao Wang
\end{minipage}
}
}

\maketitle

\begin{abstract}
In multi-source ECG deployment, new sources may arrive when earlier raw ECGs cannot be retained or replayed. Isolating source-specific classifiers on a frozen backbone prevents parameter interference, but source-unknown inference still requires selecting an appropriate expert. We study this distinction with \ours{}, a controlled continual-deployment framework built on frozen 1024-dimensional ECGFounder features. Each arriving source adds an isolated Balanced-Softmax linear expert, while a lightweight router is re-fitted using retained frozen training features and source labels from previously observed sources. Rather than proposing a new routing architecture, the main contribution is to separate preserved expert performance from autonomous source inference and quantify the resulting deployment gap. Across CPSC, PTB-XL, Georgia, and Chapman-Shaoxing, source-aware expert selection reaches $0.7915\pm0.0036$ Macro-F1, close to a matched offline independent-head reference at $0.7885\pm0.0009$. Without source IDs, an MLP router reaches $0.7756\pm0.0027$, while top-2 margin fusion reaches $0.7782\pm0.0022$. The top-2 improvement is small ($+0.0026$) and not statistically significant under paired bootstrap. Across three domain orders, the top-2-to-oracle gap remains $0.0111$--$0.0133$, indicating a persistent source-inference gap within this protocol. The results are record-level because reliable patient identifiers were unavailable. The method replays no raw ECGs, but it retains frozen feature vectors for router updates and is therefore raw-ECG-replay-free rather than memory-free. Code is publicly available at \url{https://github.com/yufanlu221/IRFE-ECG}.
\end{abstract}


\section{Introduction}

ECG systems are often deployed across changing institutions, devices, cohorts,
and annotation protocols. Continual cardiac-signal studies have considered
shifts across diseases, time, modalities, and institutions~\cite{kiyasseh2021clops},
while multicenter ECG work has highlighted practical data-sharing constraints
in sequential adaptation~\cite{kim2024multicenter_ecg_cl}. When historical raw
ECGs cannot be retained, a frozen foundation representation with lightweight
source-specific classifiers offers a practical way to add new sources without
re-optimizing a large backbone or overwriting earlier classifier parameters.

Parameter isolation, however, does not by itself solve source-unknown
deployment. If source metadata are available, the corresponding frozen expert
can be selected directly. If metadata are unavailable, the system must infer
which expert should process each record. These are different questions:
\emph{expert retention} asks whether previously learned decision boundaries are
preserved, whereas \emph{autonomous source inference} asks whether the correct
expert can be selected at test time. Conflating them can make a source-aware
expert bank appear fully solved even when autonomous deployment still incurs a
measurable gap.

This paper studies that distinction under a controlled frozen-feature protocol.
First, we instantiate \ours{}, in which each arriving ECG source adds an
isolated Balanced-Softmax linear expert on frozen ECGFounder features and
previous experts are never updated. Second, we compare several lightweight
source routers and evaluate a validation-calibrated top-2 margin rule as a
secondary operational refinement, not as the main methodological claim. Third,
we use a source-aware oracle and a matched offline independent-head reference
to separate preserved expert performance from autonomous routing performance.
The analysis shows that isolated heads remain close to their matched offline
reference, while source-unknown inference leaves a reproducible gap across
domain orders and feature-memory budgets.

\section{Related Work}

\subsection{ECG Representations and Cross-Domain Deployment}
ECG foundation models provide transferable representations for downstream
analysis. We use ECGFounder only as a frozen single-lead feature extractor and
do not fine-tune its backbone during continual learning~\cite{li2025ecgfounder}.
Cross-dataset analyses such as MELEP show that ECG representations can behave
differently across datasets and label spaces~\cite{nguyen2024melep}.
Domain-adaptation and domain-generalization methods instead seek target-adapted
or domain-invariant predictors~\cite{deng2021muda_ecg,shang2021ddg_ecg,he2023mludaf_ecg}.
Our fixed-representation setting preserves source-specific decision boundaries
and asks whether the applicable expert can be inferred without source metadata.

\subsection{Continual Learning, Isolation, and Routing}
Shared-parameter continual-learning methods mitigate interference through
regularization, distillation, or rehearsal, including EWC~\cite{kirkpatrick2017overcoming},
SI~\cite{zenke2017continual}, LwF~\cite{li2017learning}, and iCaRL
~\cite{rebuffi2017icarl}. ECG-specific studies have examined replay and
sequential adaptation across cardiac-signal or multicenter settings
~\cite{kiyasseh2021clops,kim2024multicenter_ecg_cl,gao2023ecgcl,rahmani2025dreamcl}.
Parameter-efficient approaches instead isolate prompts, adapters, or experts;
representative examples include L2P~\cite{wang2022l2p}, CODA-Prompt
~\cite{smith2023coda}, and Expert Gate~\cite{aljundi2017expertgate}. CLOM further
emphasizes that continual-learning evaluation changes when task identity is not
available at test time~\cite{kim2022clom}. These studies motivate our focus on
the distinction between retaining isolated experts and selecting them
autonomously.

\subsection{Imbalanced ECG Classification}
Normal/abnormal label frequencies differ across sources, so accuracy alone can
obscure minority-class behavior. Balanced Softmax incorporates class frequency
into the training objective~\cite{ren2020balanced}, while logit adjustment is a
related long-tail strategy~\cite{menon2021longtail}. We train each source expert
with Balanced Softmax, calibrate its decision threshold using validation data,
and use Macro-F1 as the primary classification metric.

\begin{figure*}[!t]
\centering
\includegraphics[width=1\textwidth]{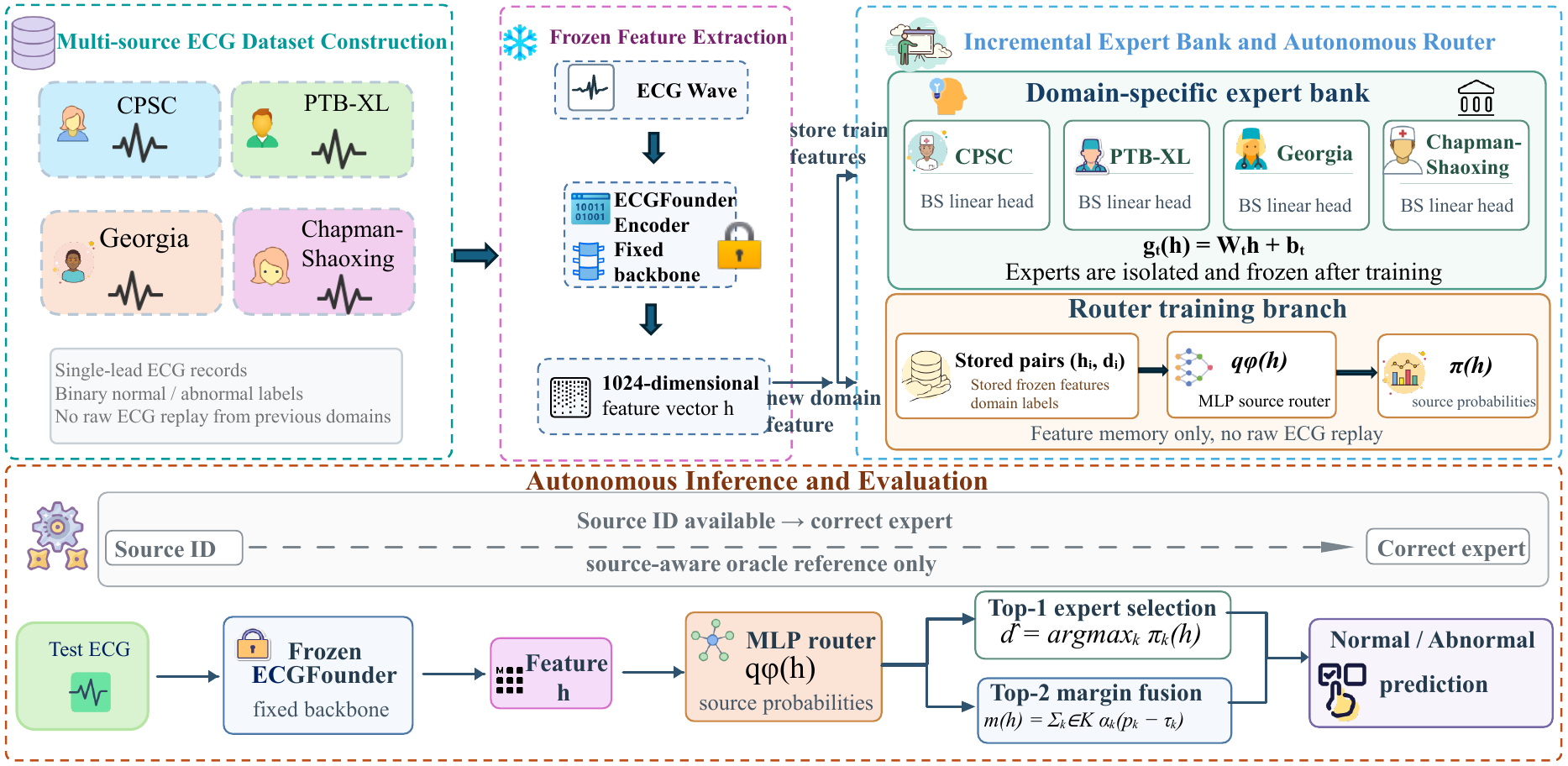}
\vspace{-1mm}
\caption{Overview of \ours{}. A frozen ECGFounder encoder maps each ECG to a
1024-dimensional feature vector. Each arriving domain adds an isolated
balanced-softmax linear expert, while previous experts and the backbone remain
frozen. Frozen train features and domain labels from seen domains are retained
to update an autonomous source router without replaying raw ECG waveforms. At
test time, source metadata are unavailable; the router selects the most likely
experts, and the top-2 variant combines validation-calibrated decision margins.
The source-aware path is an oracle reference only and is not used in autonomous
deployment.}
\vspace{-1mm}
\label{fig:framework}
\end{figure*}

\section{Method}

\subsection{Problem Setup}
Let $\mathcal{D}_1,\ldots,\mathcal{D}_T$ be a sequence of ECG domains. Domain
$\mathcal{D}_t=\{(x_i,y_i)\}_{i=1}^{n_t}$ contains single-lead ECGs $x_i$ and
binary normal/abnormal labels $y_i\in\{0,1\}$. At step $t$, the learner receives
the train and validation partitions of $\mathcal{D}_t$. Raw ECGs from
$\mathcal{D}_{1:t-1}$ are not replayed. For router updates, we retain frozen
feature vectors and domain identities from earlier training samples. The test
partitions remain inaccessible until final evaluation. Thus, the protocol is raw-ECG-replay-free, but it retains feature-level memory
for router updates and should not be interpreted as memory-free continual
learning.

We distinguish two inference settings. The \emph{source-aware reference}
provides the source ID $d$ and directly selects expert $g_d$. The primary
\emph{autonomous setting} withholds $d$ and requires a router to select or fuse
experts. This separation is important because isolated expert parameters are
not updated after their domains are learned, while autonomous predictions can
still deteriorate through routing errors.

For shared-parameter baselines, we record the conventional result matrix
$R\in\mathbb{R}^{T\times T}$, where $R_{t,j}$ is Macro-F1 on domain $j$ after
training through domain $t$, and compute
\begin{equation}
\mathrm{BWT}=\frac{1}{T-1}\sum_{j=1}^{T-1}\left(R_{T,j}-R_{j,j}\right).
\end{equation}
For the routed expert bank, however, changes in $R$ can reflect router changes
rather than expert forgetting. We therefore use routed BWT only as a diagnostic
and treat final autonomous Macro-F1 and route accuracy as the primary measures.

\subsection{Frozen Foundation Experts}
The frozen ECGFounder backbone maps an ECG to
\begin{equation}
h=f_{\theta}(x),\qquad h\in\mathbb{R}^{1024},
\end{equation}
where $\theta$ is never updated. When domain $t$ arrives, a linear expert
$g_t$ is added:
\begin{equation}
\ell_t=g_t(h)=W_t h+b_t, \qquad W_t\in\mathbb{R}^{2\times1024}.
\end{equation}
Only $g_t$ is optimized; $g_{1:t-1}$ remain frozen. Experts use Balanced
Softmax with class counts estimated from the current domain's train split.
After model selection, a domain-specific threshold $\tau_t$ is calibrated on
that domain's validation probabilities and then fixed for test evaluation.

This architecture deliberately isolates domain decision boundaries and freezes
each expert after its domain is learned. Therefore, source-aware performance is
best interpreted as an oracle control for preserved isolated-head performance,
not as evidence that test-time source inference has been solved. We also
compare against a matched offline reference that trains the same independent
heads with all domain training sets available concurrently.

\subsection{Incremental Domain Router}
At step $t$, the router is trained only over the seen domain set
$\mathcal{S}_t=\{1,\ldots,t\}$. Our discriminative router is a compact MLP with
a 1024-dimensional input, LayerNorm, a 128-unit hidden layer, GELU activation,
dropout, and a linear output layer whose dimension equals $|\mathcal{S}_t|$:
\begin{equation}
q_{\phi_t}(h)=W_2\,\mathrm{Dropout}\!\left(
\mathrm{GELU}(W_1\,\mathrm{LN}(h))\right).
\end{equation}
It is optimized by cross-entropy on stored train features and domain labels
from $\mathcal{S}_t$. The checkpoint maximizing route accuracy on concatenated
seen-domain validation features is restored. The router is re-fitted after
each newly observed domain; it never receives future-domain features or test
labels. The feature-memory budget reported later counts stored train-feature
rows used for router fitting. Validation features are used for model selection
in the experimental protocol and are not counted in the router training memory
budget.

For comparison, we evaluate a cosine nearest-centroid router, a cosine kNN
router over retained features, a shared-covariance shrinkage-LDA router, and a
linear domain classifier. The LDA router estimates one domain mean and a single
shared covariance matrix with shrinkage regularization using only seen-domain
train features. These controls test whether a single centroid, local
neighborhoods, linear separation, or a nonlinear discriminative boundary is
needed in the frozen feature space.

\subsection{Top-2 Validation-Margin Fusion}
A hard router applies only the highest-scoring expert. To reduce sensitivity
to hard top-1 expert selection, let $\mathcal{K}(h)$ contain the two experts
with the largest router logits $s_k(h)$. Their normalized weights are
\begin{equation}
\alpha_k(h)=\frac{\exp(s_k(h)/T_r)}
{\sum_{j\in\mathcal{K}(h)}\exp(s_j(h)/T_r)},
\end{equation}
where $T_r=1$ is fixed before evaluation. Expert $k$ produces positive-class
probability $p_k(h)$ and has a domain-specific validation threshold $\tau_k$.
We fuse calibrated margins rather than raw probabilities:
\begin{equation}
m(h)=\sum_{k\in\mathcal{K}(h)}\alpha_k(h)\bigl(p_k(h)-\tau_k\bigr),
\qquad \hat y=\mathbb{I}[m(h)\geq0].
\end{equation}
The rule is used only for autonomous inference and uses no source labels, test
labels, or future-domain examples at evaluation time.

\section{Experimental Setup}

\subsection{Datasets and Splits}
We use four ECG domains---CPSC, PTB-XL, Georgia, and Chapman-Shaoxing---from
the preprocessed PhysioNet/Computing in Cardiology Challenge 2021 cache used in
our experiments~\cite{reyna2021will}. We keep the original source names to
denote the domain of each record. PTB-XL and Chapman-Shaoxing are also cited
through their original dataset descriptions for provenance
~\cite{wagner2020ptbxl,zheng2020chapman}. Each ECG is represented in the cache
as a fixed-length single-lead waveform with 2,500 samples and is mapped to a
binary normal/abnormal label. The main domain order is CPSC $\rightarrow$
PTB-XL $\rightarrow$ Georgia $\rightarrow$ Chapman-Shaoxing.
We use ECGFounder as an off-the-shelf frozen feature extractor and do not
fine-tune it on any train, validation, or test waveform in this study. Because
the public pretraining composition of large ECG foundation models may overlap
with widely used ECG datasets, we do not claim de novo representation learning
or foundation-model pretraining independence from these sources. Our claims are
restricted to the downstream continual deployment protocol built on fixed
features.
\begin{table}[t]
\centering
\caption{Four-domain ECG split. Waveform-hash grouping prevents exact
train/test waveform overlap but does not guarantee patient-level isolation.}
\label{tab:data}
\small
\setlength{\tabcolsep}{5pt}
\renewcommand{\arraystretch}{1.08}
\begin{tabular*}{\columnwidth}{@{\extracolsep{\fill}}lcccc@{}}
\toprule
Domain & Train & Test & Hash overlap & Step \\
\midrule
CPSC    & 8249  & 2063 & 0 & 1 \\
PTB-XL  & 17391 & 4347 & 0 & 2 \\
Georgia & 8265  & 2066 & 0 & 3 \\
Chapman & 8188  & 2047 & 0 & 4 \\
\bottomrule
\end{tabular*}
\end{table}

We use a waveform-hash-grouped record-level split for all four domains. Records
with identical waveform hashes are assigned to the same partition, preventing
exact duplicate waveforms from crossing the train/test boundary. The processed
splits and frozen-feature caches used in this study do not preserve reliable
patient identifiers. We therefore do not claim patient-level separation. All
reported results should be interpreted as record-level benchmark evidence
rather than patient-level clinical validation.

\subsection{Training and Validation Protocol}
All formal results use seeds 42, 43, and 44. Domain experts are trained for 30
epochs with AdamW, cosine learning-rate decay, learning rate $5\times10^{-4}$,
and batch size 32. MLP and linear routers use learning rate $10^{-3}$ for at
most 50 epochs.

For each expert, checkpoints are selected by validation Macro-F1 at threshold
0.5. The selected checkpoint is then restored, and a domain-specific decision
threshold is calibrated on validation data. Router checkpoints are selected by
validation route accuracy on seen domains. All model selection, threshold
calibration, and hyperparameter choices are made without test access. The
held-out test set is evaluated once per selected model. Reported uncertainty
is the sample standard deviation across the three seeds.

\subsection{Metrics and Comparisons}
The primary metric is final Macro-F1 averaged across the four domains; route
accuracy measures source assignment. We additionally report balanced accuracy,
class-specific F1, AUROC, and AUPRC. Shared-parameter baselines use conventional
BWT, whereas routed BWT is not interpreted as expert-parameter forgetting.
Comparisons include pooled prediction; centroid, kNN, LDA, linear, and MLP
routing; source-aware and matched offline references; and shared-parameter
baselines including full fine-tuning, EWC, LwF, SI, and replay. All methods use
the same domain sequence and validation-only model selection where applicable.

\section{Results}
\label{sec:results}

\subsection{Autonomous Source Inference Leaves a Consistent Gap}
Table~\ref{tab:router} reports final four-domain Macro-F1 and route accuracy
after the last incremental step. The pooled head and centroid router are nearly
identical at $0.7551$ and $0.7557$ Macro-F1. Nonlinear MLP routing reaches
$0.7756\pm0.0027$, above centroid, kNN, LDA, and a matched linear router.
Top-2 margin fusion is numerically highest among autonomous variants at
$0.7782\pm0.0022$.

\begin{table}[t]
\centering
\caption{Final routing performance over three seeds. Source IDs are used only
by the oracle; dash indicates not applicable.}
\label{tab:router}
\footnotesize
\renewcommand{\arraystretch}{1.05}
\begin{tabular*}{\columnwidth}{@{\extracolsep{\fill}}lcc@{}}
\toprule
Selection rule & Macro-F1 & Route acc. \\
\midrule
Pooled single head & $0.7551{\pm}.0041$ & -- \\
Centroid & $0.7557{\pm}.0031$ & 0.619 \\
kNN & $0.7621{\pm}.0022$ & 0.668 \\
Shrinkage LDA & $0.7672{\pm}.0032$ & 0.777 \\
Linear router & $0.7694{\pm}.0019$ & 0.785 \\
Domain MLP & $0.7756{\pm}.0027$ & 0.815 \\
\textbf{MLP top-2 margin} & $\mathbf{0.7782{\pm}.0022}$ & 0.815 \\
MLP top-2 probability avg. & $0.7501{\pm}.0022$ & 0.815 \\
\midrule
Source-aware oracle & $0.7915{\pm}.0036$ & 1.000 \\
\bottomrule
\end{tabular*}
\end{table}

The source-aware oracle remains $0.0133$ Macro-F1 above MLP top-2, quantifying
the performance available when the true source identity is supplied. The
$+0.0026$ top-2 gain over hard MLP has a paired-bootstrap 95\% confidence
interval of $[-0.0010,0.0061]$ ($p=0.144$), so we treat it as a modest
operational refinement rather than a statistically significant improvement.
Equal-probability top-2 averaging performs substantially worse, and no
autonomous router uses future-domain features or test labels.

\subsection{Domain Boundaries and Routing Confusion}
\begin{figure*}[!t]
\centering
\includegraphics[width=0.95\textwidth]{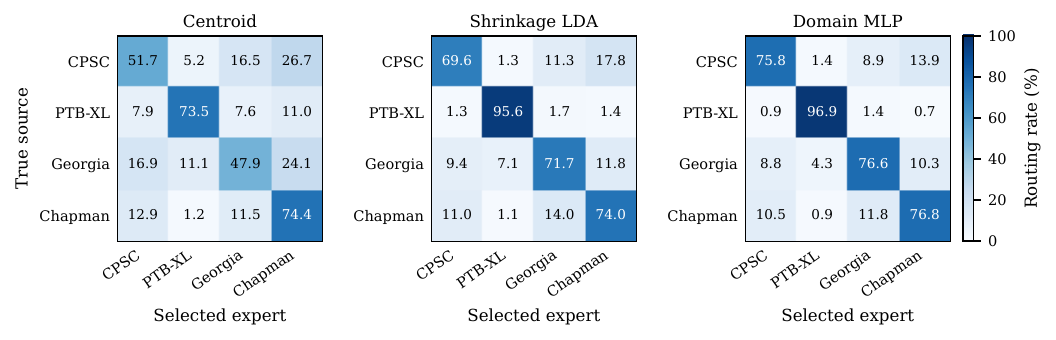}
\vspace{-1mm}
\caption{Final-step route confusion matrices averaged over seeds 42--44. Rows
are true sources and columns are selected experts; entries are percentages.}
\vspace{-1mm}
\label{fig:route_confusion}
\end{figure*}

Figure~\ref{fig:route_confusion} indicates that the improvement over centroid
routing is largely associated with better domain-discriminative boundaries. MLP
routing raises correct source selection for CPSC from 51.7\% under centroid
routing to 75.8\%, and for Georgia from 47.9\% to 76.6\%. LDA recovers much of
this gap, while the nonlinear MLP is strongest on both domains. Residual
CPSC--Chapman and Georgia--Chapman confusion explains why autonomous
performance remains below the oracle.

\subsection{Domain-Wise Effects and the Oracle Gap}
\begin{figure}[t]
\centering
\includegraphics[width=0.92\columnwidth]{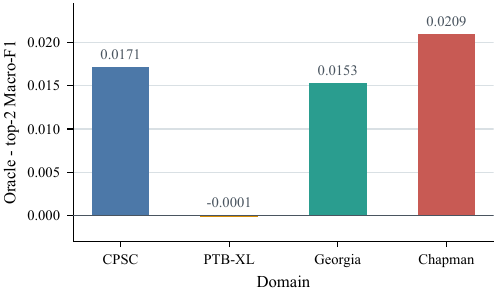}
\vspace{-1mm}
\caption{Domain-wise Macro-F1 gap between source-aware oracle selection and
autonomous MLP top-2 routing. Positive values indicate the remaining
autonomous source-inference gap.}
\vspace{-1mm}
\label{fig:domain_oracle_gap}
\end{figure}

Table~\ref{tab:domain} shows that the modest top-2 gain is concentrated in
Chapman, where Macro-F1 rises from 0.8922 to 0.9032. Changes on CPSC and
PTB-XL are small, while Georgia is slightly lower than hard MLP routing.
Figure~\ref{fig:domain_oracle_gap} visualizes the domain-wise gap between
source-aware oracle selection and autonomous MLP top-2 routing. Although CPSC
remains the lowest-performing autonomous domain, the largest top-2-to-oracle
gap occurs on Chapman, where top-2 reduces the hard-routing gap from 0.0319 to
0.0209 but does not close it.

\begin{table}[t]
\centering
\caption{Final Macro-F1 by domain, averaged over seeds 42--44.}
\label{tab:domain}
\footnotesize
\renewcommand{\arraystretch}{1.05}
\begin{tabular*}{\columnwidth}{@{\extracolsep{\fill}}lcccc@{}}
\toprule
Rule & CPSC & PTB-XL & Georgia & Chapman \\
\midrule
Oracle & 0.7412 & 0.7509 & 0.7499 & 0.9241 \\
Centroid & 0.6940 & 0.7356 & 0.7128 & 0.8804 \\
LDA & 0.7034 & 0.7510 & 0.7310 & 0.8835 \\
MLP & 0.7226 & 0.7508 & 0.7369 & 0.8922 \\
MLP top-2 & 0.7241 & 0.7510 & 0.7346 & 0.9032 \\
\bottomrule
\end{tabular*}
\end{table}

\subsection{Source-Aware and Offline References}
The matched offline independent-head reference reaches
$0.7885\pm0.0009$ Macro-F1, compared with $0.7915\pm0.0036$ for source-aware
sequential selection. Each binary linear expert contains 2,050 trainable
parameters, so the four-domain bank adds 8,200 expert parameters beyond the
frozen backbone. Their near-equality is a control for preserved isolated-head
performance: with source identity supplied, sequentially added heads behave
similarly to the same heads trained with all domains available offline. The
source-aware row is therefore an oracle reference, not an autonomous deployment
result.

Table~\ref{tab:clinical} reports additional metrics for autonomous MLP top-2
routing. The high Chapman Macro-F1 is accompanied by high balanced accuracy and
strong F1 scores for both classes, rather than being created solely by the
majority class. CPSC and Georgia retain lower negative-class F1, consistent
with the remaining ambiguity observed in the routing analysis.

\begin{table}[t]
\centering
\caption{Additional classification metrics for autonomous MLP top-2 routing,
averaged over three seeds. $F1_-$ and $F1_+$ denote F1 scores for the
negative ($y=0$) and positive ($y=1$) classes, respectively.}
\label{tab:clinical}
\small
\renewcommand{\arraystretch}{1.05}
\begin{tabular*}{\columnwidth}{@{\extracolsep{\fill}}lccccc@{}}
\toprule
Domain & BAcc & $F1_-$ & $F1_+$ & AUROC & AUPRC \\
\midrule
CPSC & 0.784 & 0.511 & 0.937 & 0.903 & 0.989 \\
PTB-XL & 0.784 & 0.677 & 0.825 & 0.863 & 0.943 \\
Georgia & 0.749 & 0.566 & 0.903 & 0.881 & 0.974 \\
Chapman & 0.903 & 0.832 & 0.974 & 0.985 & 0.998 \\
\bottomrule
\end{tabular*}
\end{table}

\subsection{Order Robustness and Feature-Memory Budget}
\begin{figure}[!t]
\centering
\includegraphics[width=0.92\columnwidth]{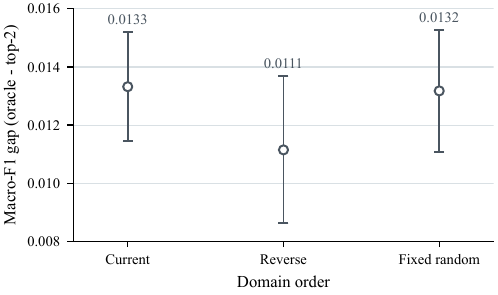}
\vspace{-1mm}
\caption{Within-order Macro-F1 gap between source-aware oracle selection and
autonomous MLP top-2 routing across three domain orders. Error bars denote
sample standard deviations over three seeds.}
\vspace{-1mm}
\label{fig:order_gap}
\end{figure}

Figure~\ref{fig:order_gap} compares top-2 with the oracle trained under the same
domain order. The gap is 0.0133 for the main order, 0.0111 for the reverse
order, and 0.0132 for a fixed random order. We emphasize within-order gaps
because changing the order also changes the random training trajectory of the
domain experts. The stable gap suggests that autonomous source inference remains a consistent
limitation rather than a peculiarity of one domain order.

\begin{figure}[!t]
\centering
\includegraphics[width=0.96\columnwidth]{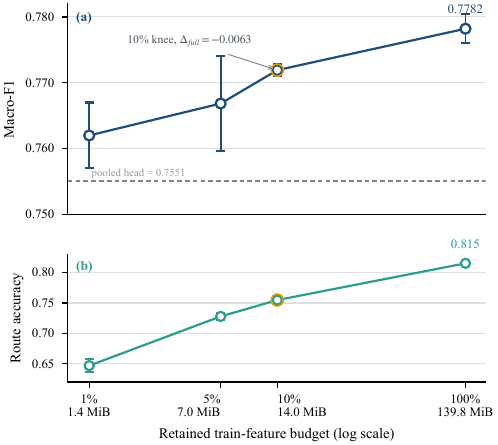}
\vspace{-1mm}
\caption{Feature-memory trade-off for autonomous MLP top-2 routing.
(a) Final Macro-F1 under different retained train-feature budgets.
(b) Test route accuracy under the same budgets. The dashed line marks the
pooled single-head baseline. Error bars denote sample standard deviations over
three seeds. The 10\% setting approaches the full-memory result while using
substantially less feature storage.}
\vspace{-1mm}
\label{fig:feature_memory_tradeoff}
\end{figure}

Figure~\ref{fig:feature_memory_tradeoff} summarizes the feature-memory
trade-off. Retaining 10\% of train features places top-2 within 0.0063
Macro-F1 of the full-memory result while reducing feature storage from 139.82
to 13.98 MiB. The 1\% setting remains above the pooled-head control but loses
route accuracy. Retained features count frozen train-feature rows available
to the router. The MiB values include only stored 1024-dimensional float32
feature vectors; labels, validation features used for benchmark model
selection, and container overhead are excluded. Thus, the protocol avoids
raw-ECG replay but remains feature-memory based rather than memory-free.

\subsection{Shared-Parameter Continual-Learning Baselines}
Table~\ref{tab:baselines} places autonomous routing beside shared-parameter
baselines. These methods use one shared predictor and address weight
interference, whereas \ours{} grows isolated experts and must infer their
applicability. The replay baseline stores raw tensors, while \ours{} stores
frozen features; the comparison therefore contrasts deployment failure modes
rather than matched memory budgets.

\begin{table}[t]
\centering
\caption{Shared-parameter baselines over three seeds. Macro-F1 is reported as
mean $\pm$ standard deviation; BWT is the mean over the three runs. Dash
indicates that BWT is not interpreted for the routed expert bank.}
\label{tab:baselines}
\footnotesize
\renewcommand{\arraystretch}{1.05}
\begin{tabular*}{\columnwidth}{@{\extracolsep{\fill}}lcc@{}}
\toprule
Method & Macro-F1 & BWT \\
\midrule
Shared linear head & $0.7378{\pm}.0048$ & -0.0647 \\
Full fine-tuning & $0.7048{\pm}.0100$ & -0.0878 \\
EWC & $0.6937{\pm}.0582$ & -0.0817 \\
LwF & $0.6403{\pm}.0194$ & -0.1080 \\
Small replay (256/domain) & $0.7140{\pm}.0158$ & -0.0956 \\
SI & $0.7217{\pm}.0225$ & -0.0454 \\
\midrule
\ours{} (autonomous) & $\mathbf{0.7782{\pm}.0022}$ & -- \\
\bottomrule
\end{tabular*}
\end{table}

The shared linear head is the strongest shared-parameter baseline, and the
regularization or replay rows still show negative BWT. For \ours{}, routed BWT
is omitted because changes mainly reflect router behavior rather than expert
parameter forgetting.

\subsection{Expert and Loss Ablations}

\begin{table*}[!t]
\centering
\caption{Oracle-selected expert and loss ablations over three seeds. $\Delta$
is relative to the linear Balanced-Softmax main expert bank. Domain columns
report Macro-F1 averaged over seeds 42--44. CE denotes cross entropy.}
\label{tab:expert_loss_ablation}
\footnotesize
\renewcommand{\arraystretch}{1.05}
\begin{tabular*}{0.95\textwidth}{@{\extracolsep{\fill}}lcccccc@{}}
\toprule
Expert/loss variant & Macro-F1 & $\Delta$ vs. main & CPSC & PTB-XL & Georgia & Chapman \\
\midrule
Linear + Balanced Softmax & $0.7915{\pm}0.0036$ & -- & 0.7412 & 0.7509 & 0.7499 & 0.9241 \\
Linear + weighted CE & $0.7862{\pm}0.0016$ & -0.0053 & 0.7288 & 0.7515 & 0.7440 & 0.9207 \\
Residual adapter + Balanced Softmax & $0.7760{\pm}0.0015$ & -0.0155 & 0.7030 & 0.7481 & 0.7350 & 0.9180 \\
Validation-selected bank & $0.7833{\pm}0.0057$ & -0.0082 & 0.7166 & 0.7487 & 0.7499 & 0.9180 \\
\bottomrule
\end{tabular*}
\end{table*}
Table~\ref{tab:expert_loss_ablation} reports oracle-selected expert and loss
ablations. The linear Balanced-Softmax expert remains strongest
($0.7915\pm0.0036$ Macro-F1). Weighted cross entropy lowers performance, and
residual adapters further reduce Macro-F1, suggesting that extra
domain-specific capacity is not automatically beneficial in the frozen
ECGFounder feature space. The validation-selected heterogeneous bank also
underperforms the main design, indicating that the source-aware result is not
produced by expert-type search, residual adapters, or loss-function tuning. We
therefore retain the simple linear Balanced-Softmax bank; for this binary
benchmark, ECGFounder features appear sufficiently separable, while extra
adapter capacity may amplify validation noise rather than provide a useful
inductive bias.

\section{Discussion}

\subsection{What the Source Router Does and Does Not Establish}
The experiments separate preserved expert decision boundaries from autonomous
expert selection. Because the backbone and earlier heads are frozen, the
source-aware result measures how well the isolated source-specific classifiers
are retained, while the gap to autonomous inference reflects the additional
cost of choosing experts without metadata. The matched offline reference
supports this interpretation by showing that sequential presentation alone
does not explain the source-aware result under independent frozen heads.

The router discriminates dataset sources rather than learning a unified
domain-invariant diagnostic representation. Acquisition properties, cohort
composition, annotation policy, and label prevalence may all contribute. In the
present expert-bank design, each source keeps its own decision boundary and the
router estimates which boundary is most applicable. Thus, high route accuracy
should not be interpreted as physiological source identification or cross-domain
clinical generalization; we ask only how closely source-unknown selection can
approach a source-aware reference for previously observed sources.

\subsection{Top-2 Fusion and the Residual Gap}
The MLP router improves source discrimination over centroid-based routing in the
frozen feature space. Top-2 fusion uses the same router and therefore does not
improve route accuracy; it only reduces the cost of some uncertain top-1
selections by combining validation-calibrated expert margins. Its average gain
is $0.0026$ and is not statistically significant, so it is a secondary
operational option rather than a central contribution. The remaining gap is
consistent across domain orders but should not be attributed to routing alone:
expert calibration, binary-label harmonization, and source-specific class
imbalance may also contribute.

\subsection{Scope and Limitations}
The benchmark lacks reliable patient identifiers. Waveform-hash grouping removes
exact duplicate waveforms across partitions but cannot rule out different
records from the same patient, so all results are record-level rather than
patient-independent clinical validation. The setting is further restricted to
four previously observed sources, binary normal/abnormal labels, and single-lead
ECGs; unknown sources, gradual domain drift, multi-label diagnosis, and
multi-lead deployment are not evaluated.

The protocol is raw-ECG-replay-free but not memory-free because router updates
retain frozen feature vectors. Such embeddings can remain privacy-sensitive and
would require appropriate governance in real deployment. In addition, memory
budgets are not fully matched across the routed expert bank and shared-parameter
baselines: the replay baseline stores raw examples, whereas \ours{} stores
features. Those comparisons therefore illustrate different deployment failure
modes rather than a storage-normalized ranking. Finally, we do not directly compare with ECG domain-adaptation or
domain-generalization methods. The conclusions are limited to the internally
matched continual expert-bank protocol studied here.

\section{Conclusion}
We separated preserved expert performance from autonomous source inference in a
raw-ECG-replay-free continual ECG benchmark. Source-aware experts reach
$0.7915\pm0.0036$ Macro-F1, close to a matched offline reference; MLP routing
reaches $0.7756\pm0.0027$, while the small top-2 increase to
$0.7782\pm0.0022$ is not statistically significant. In this seen-source,
record-level protocol, the residual deployment gap is associated primarily with
autonomous expert selection rather than parameter interference. The method
retains frozen features and is not memory-free.

\section*{Acknowledgment}
\vspace{-0.5em}
This work was supported by the Fundamental Research Funds for the Central Universities (BKSJ26011).

\bibliographystyle{IEEEtran}
\bibliography{references}

\end{document}